%% file: main.tex
\documentclass{article}
\usepackage{ijcai26}

\usepackage{times}
\usepackage{soul}
\usepackage{url}
\usepackage[hidelinks]{hyperref}
\usepackage[utf8]{inputenc}
\usepackage[small]{caption}
\usepackage{graphicx}
\usepackage{enumitem}
\usepackage{amsmath}
\usepackage{amsthm}
\usepackage{booktabs}
\usepackage[switch]{lineno}

\usepackage{xcolor}
\definecolor{algcommentgray}{gray}{0.6}

\usepackage{lipsum}

\usepackage{algorithm}
\usepackage{algpseudocode}
\algnewcommand{\LeftComment}[1]{\Statex \(\triangleright\) #1}
\algrenewcommand\algorithmiccomment[1]{%
  \hfill\textit{\textcolor{algcommentgray}{\(\triangleright\) #1}}%
}
\usepackage[T1]{fontenc}
\usepackage[utf8]{inputenc}
\usepackage{svg}
\usepackage{microtype}
\usepackage{amsmath}
\usepackage{inconsolata}
\usepackage{multirow}
\usepackage{multicol}
\usepackage{graphicx}

\usepackage{amsmath}
\DeclareMathOperator*{\argmax}{arg\,max}

\title{Automatic Prompt Optimization for Dataset-Level Feature Discovery}

\author{
Adrian Cosma$^1$ \and Oleg Szehr$^1$ \and David Kletz$^1$ \and Alessandro Antonucci$^1$\\
Olivier Pelletier$^2$\\
\affiliations
$^1$ SUPSI, Dalle Molle Institute for Artificial Intelligence Studies (IDSIA)\\
$^2$UBS Switzerland AG and its affiliates\\
\emails
{\small
\{adrian.cosma, oleg.szehr, david.kletz, alessandro.antonucci\}@supsi.ch\\
olivier.pelletier@ubs.com}
}

\begin{document}

\maketitle

\begin{abstract}
    Feature extraction from unstructured text is a critical step in many downstream classification pipelines, yet current approaches largely rely on hand-crafted prompts or fixed feature schemas. We formulate feature discovery as a dataset-level prompt optimization problem: given a labelled text corpus, the goal is to induce a global set of interpretable and discriminative feature definitions whose realizations optimize a downstream supervised learning objective. To this end, we propose a multi-agent prompt optimization framework in which language-model agents jointly propose feature definitions, extract feature values, and evaluate feature quality using dataset-level performance and interpretability feedback. Instruction prompts are iteratively refined based on this structured feedback, enabling optimization over prompts that induce shared feature sets rather than per-example predictions. This formulation departs from prior prompt optimization methods that rely on per-sample supervision and provides a principled mechanism for automatic feature discovery from unstructured text.
\end{abstract}


\section{Introduction}
\label{sec:intro}
\input{sections/1.intro}

\section{Related Work}
\label{sec:related}
\input{sections/2.related}

\section{Method}
\label{sec:method}
\input{sections/3.method}

\section{Experimental Setup}
\label{sec:exps}
\input{sections/4.experimental-setup}

\section{Results}
\label{sec:results}
\input{sections/5.results}

\section{Conclusions}
\label{sec:conclusions}
\input{sections/6.conclusions}

\section*{Acknowledgements}
\input{sections/7.ack}


\bibliographystyle{named}
\bibliography{refs}

\end{document}

%% file: sections/1.intro.tex

Contemporary business organizations across industries, including sectors such as finance~\cite{aguda2024large}, healthcare~\cite{garg2025rise}, and retail~\cite{brinkmann2024using,sinha2024pae}, are increasingly relying on data-driven decision-making to enhance operational efficiency, improve customer experiences, and drive competitive advantage. A significant portion of the valuable insights required for these decisions is often embedded in text corpora  comprised of, for example, customer reviews, support tickets, clinical notes, incident reports, and internal documentation. Converting free-form content into structured representations is a crucial step towards downstream analytics and decision-making, where recent advancements in \emph{Language Models} (LMs) have significantly enhanced the feasibility of this task~\cite{he2024annollm,gilardi2023chatgpt,törnberg2023chatgpt4outperformsexpertscrowd}. 

However, even if LMs can be successfully used to extract a \textit{known} set of features from free text, a key question remains largely unexplored: \textit{for a given downstream supervised learning (SL) task (i.e., a labelled text corpus), how can one identify which are the best features that should be extracted to optimize a chosen target metric?} We distinguish feature \emph{extraction} (extracting predefined attributes) from feature \emph{discovery} (identifying which attributes are worth extracting in the first place). This situation frequently arises in business applications, where feature extraction by LMs primarily serves to inform a downstream system for decision-making~\cite{HERHAUSEN2025115491}. Furthermore stakeholders themselves often define features based on organizational or regulatory needs. These exogenous feature definitions, driven by non-technical considerations, are typically suboptimal for SL. Motivated by these points, this work places feature discovery itself at the center of the learning process, identifying features to optimize the performance of the downstream SL system.

\begin{figure*}[hbt!]
    \centering
    \includesvg[width=0.85\textwidth]{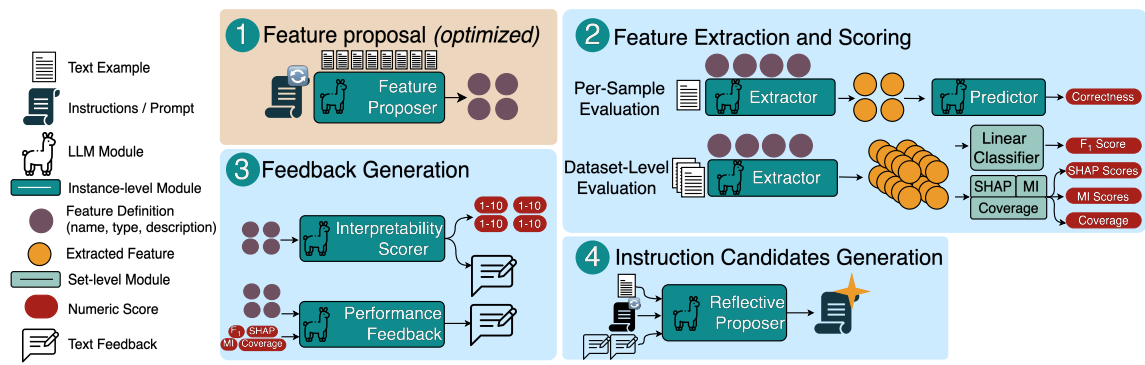}
    \caption{Diagram of proposed LM agents for instructions optimization. \textit{(1)} The \textsc{FeatureProposer} module outputs a set of feature definitions, based on the current prompt and a set of examples. \textit{(2)} Based on provided definitions, an \textsc{Extractor} is instantiated which extracts text features across the entire dataset and are evaluated. \textit{(3)} Textual feedback is gathered by \textsc{InterpretabilityScorer} and \textsc{PerformanceFeedback} modules. \textit{(4)} Based on the feedback, new instructions for the \textsc{FeatureProposer} are proposed by the \textsc{ReflectiveProposer.}}
    \label{fig:diagram}
\end{figure*}

In this work, we frame feature discovery as a prompt optimization problem~\cite{khattab2024dspy,pryzant-etal-2023-automatic,yuksekgonul2025optimizing}. 
Instead of relying on hand-crafted features or prompt heuristics, we present a procedure to refine prompts to guide LMs to propose and extract features of increasing quality. In order to propose a relevant list of domain-specific features, the model must account for the entire corpus, and since it is often impractical to include the full dataset in a prompt decisions must be based on a subset of text samples and a specific instruction set. This leads to a key question: How can one determine the most effective set of instructions that grounds the resulting feature definitions in these selected examples?

While \emph{Large} LMs (LLMs) excel at structured extraction, their computational cost makes iterative, dataset-level optimization impractical in many settings \cite{wang2025slot,gilardi2023chatgpt}. In contrast, \emph{Small} LMs (SLMs) are more appropriate for this task than LLMs due to their significantly lower computational cost, faster inference, and their ability to run locally~\cite{belcak2025small}. Building on prior work on the joint optimization of instructions and demonstrations~\cite{opsahl2024optimizing}, we introduce a prompt optimization algorithm for feature discovery. In contrast to existing prompt optimization algorithms such as MIPRO~\cite{opsahl2024optimizing} and GEPA~\cite{agrawal2025gepareflectivepromptevolution}, which optimize prompts using per-sample evaluations, our approach derives features from feedback computed at the dataset level. As such, we employ a coordinated system of LLM agents for feature discovery, as shown in Figure \ref{fig:diagram}: a \textsc{FeatureProposer} suggests feature definitions, an \textsc{Extractor} extracts features from the text given their definitions, while the \textsc{InterpretabilityScorer} evaluates the interpretability of these features. Instructions are refined by a \textsc{ReflectiveProposer} module through textual feedback from a \textsc{PerformanceFeedback} module. These mechanisms, optimized based on downstream task performance, enable the system to propose, extract, and evaluate features effectively — without relying on hand-crafted heuristics. Importantly, this agentic approach is specifically designed to empower the use of SLMs, enabling efficient feature discovery even with more limited model capacity.

In summary, this paper makes the following main contributions: 

\begin{enumerate}
    \item We frame feature discovery from unstructured text as a dataset-level prompt optimization problem, where prompts induce a shared set of features optimized for a downstream classifier rather than per-instance predictions.
    
    \item We propose a multi-agent prompt optimization algorithm for feature discovery that jointly optimizes instruction prompts and example selection using dataset-level feedback, enabling interpretable and discriminative feature discovery.

    \item We empirically show that our prompt optimization method surpasses previous LM-based feature discovery approaches that rely on hand-crafted prompts, and current prompt optimizers such as MIPRO that were designed for per-example feedback.
\end{enumerate}

%% file: sections/2.related.tex
Our approach broadly falls into the category of AutoML \cite{automl2021} as a hands-off approach to feature discovery. While the purpose of AutoML is \textit{"enabling domain experts to automatically build applications without machine learning knowledge"} \cite{zoller2021benchmark}, our approach is more related to the area of knowledge engineering \cite{allen2023knowledge}. We further contextualize our approach in terms of LLM-based feature extraction and discovery, and automatic prompt optimization. 

\subsection{Text Annotation using LMs}

The task of text annotation has been studied extensively across various disciplines, including finance~\cite{aguda2024large}, retail business~\cite{brinkmann2024using,sinha2024pae}, healthcare~\cite{bolton2024biomedlm,li2024cancerllm}, including psychology~\cite{hassan2024automated,liu2025enhanced} and telemedicine~\cite{niculae-etal-2025-dr}, and others. While some scepticism remains regarding the reliability of LLM-extracted features~\cite{reiss2023testing}, multiple studies report strong performance, often surpassing human annotators in specific contexts~\cite{he2024annollm,gilardi2023chatgpt,törnberg2023chatgpt4outperformsexpertscrowd}. For instance, ChatGPT has been shown to outperform crowd workers in certain annotation tasks~\cite{gilardi2023chatgpt}. Additionally, the development of structured generation paradigms~\cite{outlines} has further improved the reliability of LLM-generated annotations.  In our case, these results provide motivating arguments for our use of LLMs to reliably extract features from texts.

\subsection{Feature Discovery using LMs}

Recent works \cite{balek2025llm,zhou-etal-2024-llm-feature,zhang2025dynamicadaptivefeaturegeneration} showed that LLMs can propose meaningful features from unstructured text, by leveraging LLMs' linguistic understanding to identify task-specific characteristics. For example, \cite{balek2025llm} used LLMs to propose interpretable features yielding representations that enable effective rule learning. However, the authors did not use any form of feedback to inform the prompt on the resulting features and rely on hand-crafted prompts. Similarly, \cite{zhou-etal-2024-llm-feature} used LLMs to generate utterrance-level and dialogue-level features for dialogue constructiveness assessment, which are then evaluated with various statistical tools. Like \cite{balek2025llm}, their pipeline is not designed for iterative feedback on the proposed features, but rather is constructed upon a single LLM call with hand-crafted prompts. Instead, we propose a prompt optimization algorithm to iteratively discover the most appropriate prompt to generate interpretable and discriminative features. \cite{zhang2025dynamicadaptivefeaturegeneration} used a multi-agent topology to perform feature engineering on an existing feature set based on the performance on a downstream task. However, the authors focus on tabular datasets having an already existing features, whereas we are targeting unstructured data and creating features \textit{de novo}. 
Distinct from these prior approaches, our method operates on unstructured text and introduces an automatic prompt optimization algorithm that searches for the best feature set using feedback scores computed across the entire dataset.

\subsection{Prompt Optimization}

Text-based signals convey more detailed information than a single scalar reward, which can be leveraged to provide a more effective signal to the learning system, thereby enabling faster and more efficient learning by guiding agents more precisely~\cite{agrawal2025gepareflectivepromptevolution,lee2025feedback}.
Thus, \emph{Automatic Prompt Optimization} (APO) has emerged as a promising alternative to maximize task performance of instruction-tuned LMs~\cite{zhou2022large,pryzant-etal-2023-automatic,yang2023large,agrawal2025gepareflectivepromptevolution,yuksekgonul2025optimizing,xu2025metatextgrad,lee2025feedback}.
While manual prompt engineering of LMs can be a time-consuming and error prone process, the LMs' ability to understand an manipulate language enables the use of LMs themselves as prompt optimizers \cite{yang2023large}. In literature, multiple prompt optimizers have been proposed, for example ProTeGi \cite{pryzant-etal-2023-automatic}, ORPO \cite{yang2023large}, MIPRO \cite{opsahl2024optimizing}, GEPA \cite{agrawal2025gepareflectivepromptevolution}, \textit{et alia}, some implemented in frameworks such as DSPy \cite{khattab2024dspy}. Some works such as MIPRO \cite{opsahl2024optimizing} optimize both instructions and demonstrations, bootstrapping few-shot examples by uniformly selecting examples from the training set. In our formulation, we do not have "demonstrations" in the sense of inputs and outputs, but rather a set of texts from the dataset. Further, compared to other approaches, we formulate our prompt optimization procedure in terms of a dataset-level feedback signal.

%% file: sections/3.method.tex
\subsection{Problem Definition}

Our objective is to optimize the performance of an SL task, where we are given a labelled text corpus, $\mathcal{D} := \{(t_i, y_i)\}_{i=1}^{n}$, where $y_i$ is the label assigned to the text $t_i$. First, we identify a set of feature definitions, $f:=(f_1, \dots, f_k)$ that are both interpretable and discriminative with respect to the labels in $\mathcal{D}$. The number of features $k$ can either be chosen as a hyperparameter or can be automatically decided by our system.  Given these feature definitions, an extraction procedure maps each text $t_i$ to a structured representation (i.e., feature realisations) which is then used as input to a downstream SL model.

To identify $f$, the LLM must take account of an instruction $i$ and a list of examples $d:=(d_1,...,d_l) \subseteq \mathcal{D}$, where $l$ is a hyper-parameter. Formally, we write $f=\text{FP}(\phi)$, where $\text{FP}$ is a LM module taking the role of the \textsc{FeatureProposer} and $\phi:=(i,d)$ represents the prompt comprised of an instruction and examples. Importantly, the feature definitions $f$ are global objects defined for the entire corpus and not at the level of the single text $t_i$.

Further, we introduce an \textsc{Extractor} module $E(\cdot, f)$, which maps each input $t_i$ together with the feature definitions to the realized features $\hat{f}_i := E(t_i,f)$. The quality of proposed features is evaluated using a downstream global evaluation function $M$. For prompt optimization, this yields an optimization target of the form $J(\phi) := {M}(\{(\hat{f}_i, y_i)\}_{i=1}^{n})$, where we seek the best prompt $\phi^*$ such that $\phi^* := \argmax_{\phi} J(\phi)$.  This differs from standard SL in that optimization is performed over prompts that induce a global feature set, and the objective $J(\phi)$ provides only dataset-level feedback, without per-example or per-feature supervision.

\subsection{The Optimization Process}
Current prompt optimizers such as MIPRO \cite{opsahl2024optimizing} or GEPA \cite{agrawal2025gepareflectivepromptevolution} that are currently implemented within the DSPy framework \cite{khattab2024dspy}, expect that each module obtains a separate feedback for each prediction (one-to-one feedback). We propose, instead, a dataset-level feedback setup in which feedback is provided on a value computed across a set of elements -- in our case, the F$_1$ score of a linear classifier, alongside feature importance scores obtained through SHAP \cite{lundberg2017unified}, mutual information and feature coverage scores.\footnote{Note that the use of an external computation through training the linear model / obtaining SHAP scores, aligns with the "executor" block of the topology primitives proposed by \cite{zhou2025multi}.} Such an evaluation procedure is outlined in Algorithm \ref{alg:evaluation}. Algorithm \ref{alg:the-algorithm} provides a high-level overview of our proposed prompt optimization approach. 

Overall, we can summarize our algorithm as follows:
\begin{enumerate}
    \item Gather several examples sets by uniformly sampling texts from $\mathcal{D}$ to allow the model to ground its proposals into concrete examples.
    \item Propose several sets of instructions using a \textsc{ReflectiveProposer} that incorporates text feedback on interpretability and feature set performance from an initial prompt.
    \item Find the best combination of example sets and instructions using Bayesian optimization \cite{optuna}, scoring each resulting feature definition on a dedicated "annotation" dataset $\mathcal{D}_{\text{Annotation}}$.
\end{enumerate}

The algorithm, conceptually similar to MIPRO \cite{opsahl2024optimizing} as it systematically searching for the best combination of instructions and example sets, requires having additional modules to account for set-level feedback. We further provide additional details into each LM module.

\subsection{Modules Used for Prompt Optimization}

\noindent \textbf{\textsc{FeatureProposer}}. This is the main module that follows the instructions and examples in the prompt and outputs a list of feature definitions, consisting of a snake\_case name, a Python type (\textit{bool}, \textit{int}, \textit{float} or \textit{Literal} with associated values), a description and an extraction prompt that is utilized by the \textsc{Extractor} module. The prompt of this module is the only one being optimized.

\noindent \textbf{\textsc{Extractor}}. This module is instantiated based on the feature definitions returned by the \textsc{FeatureProposer} and is tasked to extract feature values from an input text. This module is part of the evaluation.

\noindent \textbf{\textsc{InterpretabilityScorer}}.
This module assigns an interpretability score to each feature \cite{zytek2022need,molnar2020interpretable}. In particular, we follow \cite{zytek2022need}, which identified several characteristics of interpretable features: a feature needs to be readable, human-worded, understandable, meaningful and trackable. 

One potential caveat of our optimization setup is that one of the proposed features could simply be another way of expressing the target label, essentially offloading the classification to the \textsc{Extractor} module instead of the linear classifier. For example, in a text sentiment classification scenario, one of the proposed features could be "\textit{overall\_sentiment}$_{\text{categorical}}$", which, when extracted by the \textsc{Extractor} module, will result in high accuracy by the linear classifier. This is, of course, undesirable as this feature does not provide any useful structured information of the underlying texts. Consequently, this module also severely penalizes features which are leaking the label in the way we described above. This aspect is not addressed in previous works for automatic feature discovery \cite{balek2025llm}. The module is also instructed to assign low scores to unclear features names or opaque descriptions and assigning higher scores to features grounded in the source text. Furthermore, this module also outputs an additional text feedback on the interpretability of the whole feature set used later on by the \textsc{ReflectiveProposer} (detailed below). We use the interpretability score as additional "regularization" on the feature set and combine it with the performance score through a hyperparameter $\lambda$. 

\noindent \textbf{\textsc{PerformanceFeedback}}. This module provides textual feedback on the overall classifier performance of the proposed feature set. As opposed to the \textsc{InterpretabilityScorer}, this module operates on dataset-level measures (e.g., the F$_1$ score, SHAP scores).

\noindent \textbf{\textsc{ReflectiveProposer}}. This module proposes a set of instructions based on a summary of the data and a summary of the current instructions, text feedbacks from the \textsc{InterpretabilityScorer} and the \textsc{PerformanceFeedback} modules to generate instruction suggestions - similar in spirit to GEPA \cite{agrawal2025gepareflectivepromptevolution} and TextGrad \cite{yuksekgonul2025optimizing}. A version without incorporating text feedback corresponds to the \textsc{GroundedProposer} implemented in DSPy \cite{khattab2024dspy} / MIPRO \cite{opsahl2024optimizing}. Note that this step can be performed for multiple iterations, obtaining more refined prompts across multiple feedback rounds.

\input{algorithms/evaluate}
\input{algorithms/optimization}

\subsection{Overview}

Our proposed modules are designed to capture the essential aspects of a downstream classification task. Following \cite{zhou2025multi}, at each stage of optimization process additional helper modules can be incorporated that can enhance the quality of the feedback and prompt proposals without fundamentally changing the agent topology. For example performing multiple rollouts, summarizing, reflection or multi-agent debate \cite{chern2024can}. Furthermore, while the we optimize the instructions and input examples only for the \textsc{FeatureProposer} module, the other helper modules have their prompts fixed. Presumably, their instructions could also be optimized using, for example, the approach from \cite{xu2025metatextgrad}.

\subsection{Asymptotic Analysis of the Language-based Program}

We analyze our multi-agent prompt optimizer as a \emph{Language-based Algorithm} (LbA), as proposed by \cite{meyerson2025position}. In this lens, we derive the asymptotic analysis in terms of LM primitives, where the fundamental unit of cost is an \emph{LM primitive}: Following \cite{meyerson2025position}, an LM primitive is a forward-pass token generation whose cost scales as $mn$ for model size $m$ and input length $n$. 

For each candidate prompt $\phi$, the system performs a single \textsc{FeatureProposer} call to produce a global feature set $f$, $N_A=|\mathcal{D}_{\text{Annotation}}|$ calls to the \textsc{Extractor} to realize feature vectors across the annotation split, and $\mathcal{O}(1)$ additional LLM calls to \textsc{PerformanceFeedback} and \textsc{InterpretabilityScorer} to produce dataset-level textual feedback and an interpretability (Algorithms \ref{alg:evaluation} and \ref{alg:the-algorithm}). Let $L_t$ denote the average token length of an input text and $L_f$ the token length of the serialized feature schema; and let $m_{\text{FP}}, m_E, m_S$ be the model sizes used for proposing, extracting, and scoring, respectively. Under the primitive cost model, one evaluation has cost: 

\begin{equation}
C_{\text{eval}}
:=  \mathcal{O}(m_{\text{FP}} L_\phi) +  \mathcal{O}(N_A\, m_E (L_t + L_f)) +  \mathcal{O}(m_S L_f)
\end{equation}

where the second term dominates. We discard additional costs of the non-LM "executor" computations (i.e., training the classifier and computing SHAP / MI / coverage scores), since costs are dominated by the LM inference. Since Algorithm \ref{alg:evaluation} evaluates $\Theta(N_d + N_{\text{iter}})$ candidate (instruction, example sets) pairs, the total optimization cost is $ \mathcal{O}((N_d + N_{\text{iter}})\,C_{\text{eval}})$. Consequently, the computational cost is asymptotically dominated by:

\begin{equation}
  \mathcal{O}((N_d + N_{\text{iter}})\,N_A\,m_E(L_t+L_f))\,. 
\end{equation}

This highlights the fact that the end-to-end cost is primarily governed by extraction throughput over $\mathcal{D}_{\text{Annotation}}$, suggesting that practical scaling should preferentially minimize the \textsc{Extractor} primitive via smaller specialist models, while reserving larger-capability models for feature proposal and reflection.

%% file: algorithms/evaluate.tex
\begin{algorithm}[hbt!]
    \caption{Evaluation of feature definitions.} 
    \label{alg:evaluation}
    \begin{algorithmic}
     \scriptsize
        \Require $\mathcal{D}_{\text{Annotation}} = \{(t_i, y_i)\}$ - the annotation dataset; $f$ - proposed feature definitions; \textsc{Extractor}, \textsc{InterpretabilityScorer}, \textsc{PerformanceFeedback} modules.

        \Function{evaluate}{$\mathcal{D}_{\text{Annotation}}$, $f$}
        \For{$i$ in |$\mathcal{D}_{\text{Annotation}}$|}
            \State $z^{(i)} \gets \text{\textsc{Extractor}}(t_i, f)$
            \Comment{Extract features from each example $t_i$}
        \EndFor
        \State $Z := \{(z^{(i)}, y^{(i)})_{i=1}^{|\mathcal{D}_{\text{Annotation}}|}\}$
        \State Split annotated dataset intro training and validation $Z_{\text{train}}$, $Z_{\text{val}}$.
        \State Train a simple linear classifier on the dataset $Z_{\text{train}}$.
        \State Compute $F_1$ score on $Z_{\text{val}}$
        \State Compute mutual information (MI), SHAP scores and coverage on $Z_{\text{val}}$.
        \State $\tau_p \gets \text{\textsc{PerformanceFeedback}}(F_1, \text{SHAP}, \text{MI}, \text{Coverage})$
        \Comment{Get text feedback on the performance with the computed scores.}
        \State $s_I, \tau_I \gets \text{\textsc{InterpretabilityScorer}}(F_{\phi})$ 
        \Comment{Get an interpretability score per feature and text feedback for interpretability.}
        \State \Return ($\lambda F_1 + (1-\lambda)s_I$, $\tau_I + \tau_p$)
        \Comment{Return the combined score and textual feedback.}
        \EndFunction
    \end{algorithmic}
\end{algorithm}

%% file: algorithms/optimization.tex
\begin{algorithm}[hbt!]
    \caption{Finding the best instructions and sets of examples for downstream classification.} 
    \label{alg:the-algorithm}
    \begin{algorithmic}
     \scriptsize
        \Require $\mathcal{D}_{\text{Train}}$- the training dataset of texts with their classification label; $\mathcal{D}_{\text{Annotation}}$ - the annotation dataset; $\Pi_{\phi}$ the \textsc{FeatureProposer} module with initial prompt $\phi$; $N_d$ number of example sets; $l$ - number of examples per example set; $N_{\text{iter}}$ - number of optimization iterations; $N_{\text{fb}}$ - number of iterative prompt refinements. 

        \LeftComment{\textbf{Step 1: Sample example sets and generate instruction proposals}}
        \State Uniformly sample $N_d$ example sets $\{(d^1_1, \dots, d^1_{l}), (d^{N_d}_1, \dots, d^{N_d}_{l}) \}$ from $\mathcal{D}_{\text{Train}}$.

        $\mathcal{I} \gets \emptyset$, $\phi_0 \gets \phi$
        \For{$i=1$ to $N_{d}$}
                \Comment{Evaluate each example set $D$ to get scores $s_{i}$ and text feedbacks $\tau_{i}$ using Algorithm \ref{alg:evaluation}.}
            \For{$r = 1$ to $N_{\text{fb}}$}
            \Comment{Iteratively refine instructions}
                \State $\phi_r \gets (\phi_{r-1}, (d^i_1, \dots, d^i_{l}))$
                \State $f \gets \text{FP}(\phi_r)$
                \State $s_i, \tau_i \gets \text{EVALUATE}(\mathcal{D}_{\text{Annotation}}, f)$
                \State $\phi_r \gets \text{\textsc{ReflectiveProposer}}(\phi_r, D_i, s_{i}, \tau_{i})$
                \Comment{Generate a set of instruction candidates based on the current prompt, sampled examples set, and scores and feedbacks.}
            \EndFor
            \State $\mathcal{I} \gets \mathcal{I} \cup \phi_{N_{\text{fb}}}$
        \EndFor

        \LeftComment{\textbf{Step 2: Bayesian optimization over instructions and examples sets to find the best combination.}}
        
        \State $s_{\text{best}} \gets -\infty$, $\phi_{\text{best}} \gets \emptyset$
        \State Initialize Bayesian-optimization–based categorical distribution over $(I, d)$ pairs.
        \For{$k = 1$ to $N_{\text{iter}}$}
            \State Sample instruction $I_k \in \mathcal{I}$ and example set $D_k$.
            \Comment{Categorical sampling with probabilities updated by Bayesian optimization over past scores.}
            \State $\phi_k \gets (I_k, D_k)$
            \Comment{Build prompt with instructions and example sets}
            \State $f \gets \text{FP}({\phi_k})$
            \Comment{Propose features under this prompt.}
            \State $s_k \gets \text{EVALUATE}(\mathcal{D}_{\text{Annotation}}, f)$
            \If{$s_k > s_{\text{best}}$}
                \State $s_{\text{best}} \gets s_k$, $\phi_{\text{best}} \gets \phi_k$
            \EndIf
        \EndFor
        \State \Return $\phi_{\text{best}}$, $s_{\text{best}}$
    \end{algorithmic}
\end{algorithm}

%% file: sections/4.experimental-setup.tex
\paragraph{Backbone SLMs.}
We experiment with several popular open-weight SLMs, covering three model families and four scales: Qwen3-4B and Qwen3-14B \cite{yang2025qwen3}, Llama-8B \cite{grattafiori2024llama} and Gemma3-27B \cite{team2025gemma}. We use Chain-of-Thought prompting \cite{kojima2022large} for all modules except for the \textsc{Extractor} module. We use the same underlying LM as the backbone for all modules, a common scenario when computational resources are limited and do not allow multiple vLLM servers to operate in parallel. 

\paragraph{Datasets.} We use two financial news datasets annotated with sentiments from Twitter\footnote{\url{hf.co/datasets/zeroshot/twitter-financial-news-sentiment}, Accessed 19 Jan., 2026} (\textit{bearish}, \textit{neutral}, \textit{bullish} classes) and Yahoo Finance\footnote{\url{hf.co/datasets/ugursa/Yahoo-Finance-News-Sentences}, Accessed 19 Jan., 2026.} (\textit{positive}, \textit{neutral}, \textit{negative} classes). Sentiment classification from financial text is a well-established SL benchmark that requires grounding predictions in fine-grained lexical cues and domain-specific jargon. This makes it a suitable test-bed for feature discovery and extraction methods that aim to identify interpretable, label-discriminative textual properties. Generic sentiment datasets such as IMDB reviews~\cite{imdb} are not well suited to our setting, as sentiment is largely expressed through explicit polarity words, limiting the need for discovering richer, domain-specific features.

We additionally evaluate on ToxicChat~\cite{lin2023toxicchat}, a benchmark for toxicity detection in real-world user-assistant conversations. This task reflects a core challenge in contemporary LM-based systems, where moderation and safety are essential for deployment. ToxicChat exhibits linguistic properties distinct from social-media toxicity datasets, including implicit and adversarial prompts, making it a complementary setting to assess the performance of our procedure. We used the "input-only" setup of the toxicity subset: the model receives as input only the prompt and not the resulting assistant output and is tasked to determine whether the prompt is \textit{toxic} or \textit{non-toxic}.

\paragraph{Implementation Details.} We use DSPy to implement the optimization algorithm in the framework's formalisms, and \textit{optuna} \cite{optuna} for Bayesian optimization of the best combination of prompt and example sets. Compared to other approaches \cite{balek2025llm,zhang2025dynamicadaptivefeaturegeneration}, we use structured generation \cite{outlines} at each step of the pipeline. We use a temperature of 0.75 and nucleus sampling probability \cite{holtzman2019curious} of 0.95 for the \textsc{FeatureProposer} module, and greedy sampling for all other modules. Context length of the models is set to 16'000 tokens. Experiments are performed using a vLLM server \cite{kwon2023efficient} on a cluster 2xA100 GPUs. 

We use a training set $\mathcal{D}_{\text{Train}}$ consisting of 16 examples per class, from which examples are sampled, an annotation set $\mathcal{D}_{\text{Annotation}}$ consisting of 512 total examples. These sets are fixed for all experiments. Unless otherwise specified, we used $N_d=16$ examples sets; note that these examples are not LM "demonstrations" \cite{opsahl2024optimizing} which include inputs and desired program outputs, but simply a set of texts and their classification label from the dataset on which to ground feature definitions. Each example set consists of $l=16$ examples, and we set the number of feedback rounds $N_{fb}=1$, by default. The \textsc{FeatureProposal} module is initially prompted to output between 5 and 10 features. We use the $F_1$ score of a logistic regression model trained on the extracted features for performance measurement. We use $\lambda=0.75$ to weigh the contribution of interpretability estimates in the final scores during optimization.

%% file: sections/5.results.tex
\begin{figure*}[hbt!]
    \centering
    \includesvg[width=0.85\linewidth]{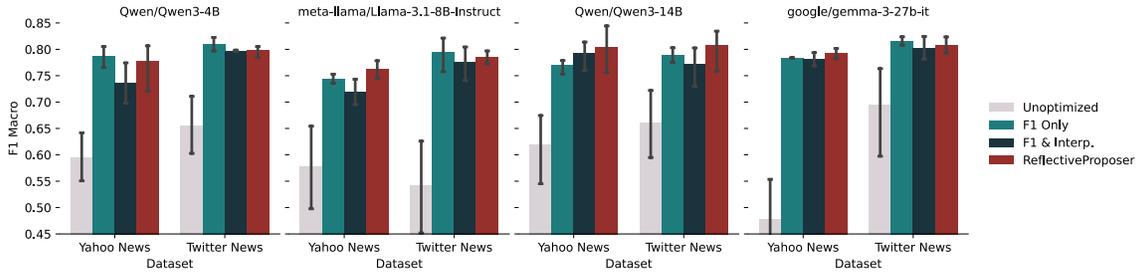}
    \caption{Comparison between \textsc{ReflectiveProposer} and variants using scalar feedback, across Yahoo and Twitter News, for four pretrained SLM models. Optimizing for F1 scores only can lead to feature leakage, artificially increasing performance, while regularizing through interpretability feedback can result in more grounded features.}
    \label{fig:llms-f1}
\end{figure*}

\paragraph{Main Results.}

In Figure \ref{fig:llms-f1} we compare our \textsc{ReflectiveProposer} which uses text feedback for refinement with variants using scalar feedback, which amounts to rephrasing of the original prompt and inclusion of general domain-specific insights. The \textsc{ReflectiveProposer} offers more specific feedback based on the global performance of resulting feature set. The "unoptimized" version roughly corresponds to the hand-crafted prompt from \cite{balek2025llm}, which is the initial seed prompt for the \textsc{FeatureProposer}. For some models, the inclusion of interpretability estimates sometimes reduces performance, since optimizing only for the final F1 score can allow the model to "cheat" by proposing features that are the same as the target label. As such, in some cases, the high performance of the F1-only scoring is artificial and due to label leakage. For example, on Yahoo and Twitter, common such "leaky" features are \textit{sentiment\_label}$_{\text{categorical}}$ or \textit{label\_sentiment\_type}$_{\text{categorical}}$ which prompt the \textsc{Extractor} to directly classify the text. In contrast, our the addition of the \textsc{ReflectiveProposer} improves upon F1 and interpretability resulting in more grounded features. Examples of proposed features include: \textit{economic\_event}$_{\text{categorical}}$, \textit{job\_loss\_indicator}$_{\text{Bool}}$, \textit{certainty\_measure}$_{\text{categorical}}$, \textit{authority\_statement}$_{\text{Bool}}$, \textit{comparative\_language\_usage}$_{\text{Bool}}$, \textit{corporate\_announcement\_type}$_{\text{categorical}}$, \textit{risk\_assessment\_tone}$_\text{categorical}$.

\begin{figure}[hbt!]
    \centering
    \includesvg[width=0.85\linewidth]{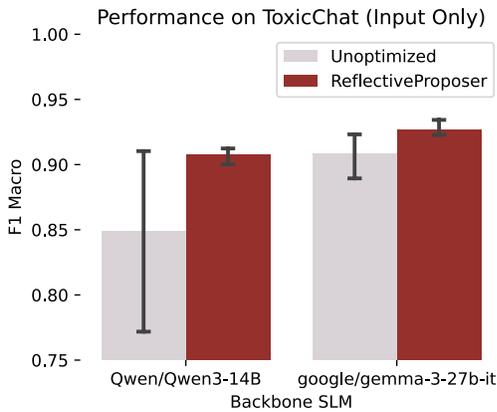}
    \caption{Evaluation on ToxicChat dataset, considering the input-only toxicity subset.}
    \label{fig:toxicchat}
\end{figure}

In Figure \ref{fig:toxicchat} we show results on the ToxicChat dataset for two backbone SLMs, comparing the unoptimized prompt with our variant, the latter resulting in consistently more discriminative features. Examples of proposed features by our system for ToxicChat are: \textit{roleplay\_instruction}$_{\text{Bool}}$, \textit{manipulative\_language}$_{\text{Bool}}$, \textit{sexual\_content\_present}$_{\text{Bool}}$, \textit{profanity\_frequency}$_{\text{float}}$, while "leaky" features that were filtered are: \textit{topic\_sensitivity}$_{\text{categorical}}$, \textit{safety\_risk\_score}$_{\text{float}}$, among others. Readers are referred to the Supplementary Materials for initial agent signatures, additional examples of proposed features, and examples of optimized prompts.

\input{tables/main-results}

\paragraph{Comparison with Per-Example Feedback.}
To make our setup compatible with a per-example feedback mechanism, we introduce a \textsc{Predictor} LM module which simply takes as input the feature values (without the original input text) and is tasked to directly classify the instance in a zero-shot manner, with no information about feature values for other examples in the dataset (see Figure \ref{fig:diagram}, panel 2). While it is clearly not ideal, it enables us to directly use already available prompt optimizers from DSPy, namely MIPRO \cite{opsahl2024optimizing}. We emphasize that optimizers such as MIPRO are not appropriate for this setting for the following reasons: \textit{(i)} there is no concept of "example set" and thus each input text will result in a different feature set; \textit{(ii)} in-context demonstrations implicitly require an input and its desired output, and thus demonstrations will contain texts and a different feature set per text; \textit{(iii)} the evaluation signal during optimization is an approximation of the true evaluation function $J$ through the \textsc{Predictor} module, which does not account for global feature realisations. Consequently, retrofitting such existing optimizers for this task is not appropriate. In Table \ref{tab:results} we show the comparison between MIPRO and our dataset-level optimization procedure using Qwen-14B as a backbone LM for ToxicChat. 
 
\begin{figure}
    \centering
    \includesvg[width=0.85\linewidth]{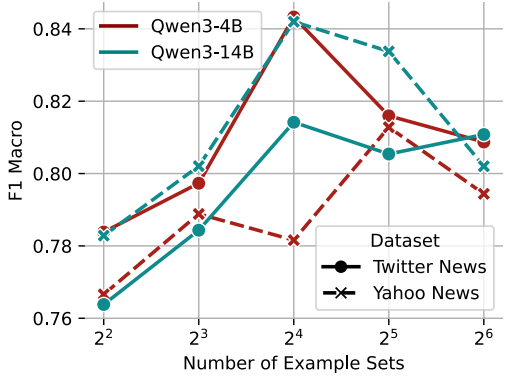}
    \caption{Evolution of accuracy by increasing number of examples sets, and, in turn, of prompt proposals, for Qwen-4B and Qwen-14B, across Yahoo News and Twitter News datasets.}
    \label{fig:iteration-steps}
\end{figure}

\paragraph{The Effect of Increasing Number of Example Sets.}
In Figure \ref{fig:iteration-steps} we show the evolution of performance when increasing the number of examples sets $N_d$, thereby increasing the search space. We heuristically set $N_{\text{iter}} = \max(N_d^2, 128)$. Results show that increasing the number of example sets generally improves performance, particularly for smaller values of $N_d$. With few example sets, prompt proposals are strongly influenced by the specific sampled texts, leading to higher variability in the induced feature sets. As $N_d$ increases, the \textsc{FeatureProposer} is exposed to a broader range of examples, resulting in more stable and discriminative features. However, gains diminish for larger values of $N_d$, and performance can slightly degrade when the search space grows faster than the available optimization budget. This suggests a trade-off between dataset coverage and effective exploration of the instruction–example space.

\begin{figure}[hbt!]
    \centering
    \includesvg[width=0.85\linewidth]{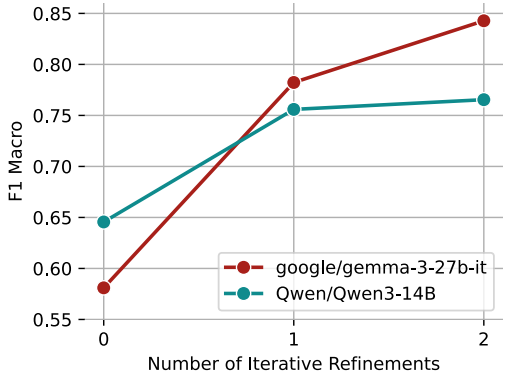}
    \caption{Performance evolution with increasing the number of iterative prompt refinements using the \textsc{ReflectiveProposer} on the Yahoo News dataset.}
    \label{fig:n-ref}
\end{figure}

\paragraph{Iterative Refinement with \textsc{ReflectiveProposer}.}
Figure \ref{fig:n-ref} shows the effect of applying multiple rounds of instruction refinement using the \textsc{ReflectiveProposer}. Performance improves with additional refinement rounds, with most of the gain occurring after the first iteration. Inspecting the prompts reveals qualitative differences: the first refinement mainly adds coarse, domain-level constraints that are absent from the initial prompt, while later refinements become increasingly specific about what constitutes a useful feature and what should be avoided. In particular, later iterations more explicitly discourage overly abstract features and instead emphasize concrete, text-grounded properties.

\paragraph{Discussion.}
We also experimented with other smaller models such as Llama3-1B and Gemma3-4B. However, they exhibited unreliable adherence to structured output schemas, frequently producing repetitive and invalid generations (e.g., repeating indefinitely a space character and not outputing valid JSON). To mitigate this, we resorted to the BAML format \cite{baml}; however, this workaround does not guarantee strict schema compliance and is therefore suboptimal. When evaluating LLama3-1B, we found it to be poorly suited for prompt optimization. The generated prompt refinements were typically generic and weakly grounded in the task, for example producing instructions such as "Generate a set of features to predict company names" or broadly phrased sentiment analysis instructions. As a result, the induced feature sets were low quality and incoherent. More generally, smaller models exhibited a higher tendency toward hallucinated or underspecified instructions and struggled both to propose meaningful feature definitions and to reliably follow refined prompts. This behaviour highlights a fundamental limitation of prompt optimization in this setting: its effectiveness is strongly constrained by the instruction-following capability of the underlying LM, which, in practice, correlates with model scale.

%% file: tables/main-results.tex
\begin{table*}[hbt!]
\centering
\caption{Comparison with MIPRO, which uses per-example feedback through the \textsc{Predictor} module, on the ToxicChat dataset, having Qwen3-14B as the underlying LM. Using per-example feedback worsens results compared to the unoptimized version.}
\label{tab:macro_metrics}
\resizebox{0.75\linewidth}{!}{\begin{tabular}{llccc}
\textbf{Method} & \textbf{Feedback} & \textbf{F1} & \textbf{Precision} & \textbf{Recall} \\
\midrule
Unoptimized & -- & 0.849 $\pm$ 0.071 & 0.869 $\pm$ 0.066 & 0.844 $\pm$ 0.07 \\
\midrule
\multirow{2}{*}{MIPRO~\cite{opsahl2024optimizing}} & \textsc{Predictor}  & 0.75 $\pm$ 0.039 & 0.819 $\pm$ 0.02 & 0.749 $\pm$ 0.034 \\
 & \textsc{Predictor} \& Interp. & 0.796 $\pm$ 0.027 & 0.844 $\pm$ 0.015 & 0.791 $\pm$ 0.026 \\
\midrule
\textbf{Ours} & \textbf{\textsc{ReflectiveProposer}} & \textbf{0.907} $\pm$ 0.006 & \textbf{0.916} $\pm$ 0.014 & \textbf{0.905} $\pm$ 0.009 \\
\end{tabular}}
\label{tab:results}
\end{table*}

%% file: sections/6.conclusions.tex
We presented a dataset-level formulation of automatic feature discovery from unstructured text, framing the problem as prompt optimization over instructions that induce global feature definitions rather than per-instance predictions. Our multi-agent framework jointly proposes feature definitions, extracts feature realizations, and refines instructions using dataset-level feedback combining downstream performance and interpretability estimates. 

Across multiple datasets and model families, we show that dataset-level textual feedback, when incorporated via our \textsc{ReflectiveProposer}, consistently improves both classification performance and feature interpretability compared to unoptimized prompts and scalar-only feedback. Interpretability regularization mitigates degenerate solutions such as label leakage, encouraging features that capture meaningful, task-relevant structure rather than directly encoding the target label.

The proposed formulation naturally applies to tasks where predictions are derived from collections of texts rather than individual instances, such as conversation-level analysis \cite{tigunova2019listening} and across social-media posts \cite{bucur2023s}. More broadly, our results suggest that prompt optimization can serve as a practical mechanism for automated, interpretable feature engineering.

The approach has two main limitations. First, dataset-level optimization is computationally expensive, as each iteration requires feature extraction over a substantial portion of the corpus. Second, the effectiveness of the method is strongly constrained by the instruction-following capabilities of the underlying language model: weaker models often fail to generate meaningful refinements. Addressing these limitations through model specialization and improved instruction-following remains an important direction for future work, and in general, for automatic prompt optimization approaches.

%% file: sections/7.ack.tex
This work was supported by UBS Switzerland AG and its affiliates through the UBS-IDSIA AI Lab initiative, where Stefania Stan and extended teams are acknowledged for funding governance and oversight, ensuring alignment with societal and industry objectives.

%% file: refs.bib
@misc{agrawal2025gepareflectivepromptevolution,
  title        = {{GEPA: Reflective Prompt Evolution Can Outperform Reinforcement Learning}},
  author       = {Lakshya A Agrawal and Shangyin Tan and Dilara Soylu and others},
  year         = {2025},
  url          = {https://arxiv.org/abs/2507.19457},
  eprint       = {2507.19457},
  archiveprefix = {arXiv},
  primaryclass = {cs.CL}
}

@inproceedings{aguda2024large,
  title        = {{Large language models as financial data annotators: A study on effectiveness and efficiency}},
  author       = {Aguda, Toyin D and Siddagangappa, Suchetha and Kochkina, Elena and others},
  year         = {2024},
  booktitle    = {Proceedings of the 2024 Joint International Conference on Computational Linguistics, Language Resources and Evaluation (LREC-COLING 2024)},
  pages        = {10124--10145}
}

@article{allen2023knowledge,
  title={Knowledge Engineering Using Large Language Models},
  author={Allen, Bradley P and Stork, Lise and Groth, Paul},
  journal={Transactions on Graph Data and Knowledge},
  volume={1},
  number={1},
  pages={3--1},
  year={2023},
  publisher={Schloss Dagstuhl--Leibniz-Zentrum f{\"u}r Informatik}
}

@article{automl2021,
  title        = {{AutoML: A survey of the state-of-the-art}},
  author       = {Xin He and Kaiyong Zhao and Xiaowen Chu},
  year         = {2021},
  journal      = {Knowledge-Based Systems},
  volume       = {212},
  pages        = {106622},
  doi          = {https://doi.org/10.1016/j.knosys.2020.106622},
  issn         = {0950-7051},
  url          = {https://www.sciencedirect.com/science/article/pii/S0950705120307516}
}

@article{balek2025llm,
  title        = {{LLM-based feature generation from text for interpretable machine learning}},
  author       = {Balek, Vojt{\v{e}}ch and S{\`y}kora, Luk{\'a}{\v{s}} and Sklen{\'a}k, Vil{\'e}m and others},
  year         = {2025},
  journal      = {Machine Learning},
  publisher    = {Springer},
  volume       = {114},
  number       = {11},
  pages        = {1--30}
}

@misc{baml,
  title        = {{BAML}},
  author       = {Boundary ML},
  year         = {2024},
  url          = {https://github.com/boundaryml/baml}
}

@article{belcak2025small,
  title        = {{Small Language Models are the Future of Agentic AI}},
  author       = {Belcak, Peter and Heinrich, Greg and Diao, Shizhe and others},
  year         = {2025},
  journal      = {arXiv preprint arXiv:2506.02153}
}

@article{bolton2024biomedlm,
  title        = {{BioMedLM: A 2.7B parameter language model trained on biomedical text}},
  author       = {Bolton, Elliot and Venigalla, Abhinav and Yasunaga, Michihiro and others},
  year         = {2024},
  journal      = {arXiv preprint arXiv:2403.18421}
}

@inproceedings{brinkmann2024using,
  title        = {{Using LLMs for the extraction and normalization of product attribute values}},
  author       = {Brinkmann, Alexander and Baumann, Nick and Bizer, Christian},
  year         = {2024},
  booktitle    = {European Conference on Advances in Databases and Information Systems},
  pages        = {217--230},
  organization = {Springer}
}

@inproceedings{bucur2023s,
  title        = {{It’s just a matter of time: Detecting depression with time-enriched multimodal transformers}},
  author       = {Bucur, Ana-Maria and Cosma, Adrian and Rosso, Paolo and others},
  year         = {2023},
  booktitle    = {European conference on information retrieval},
  pages        = {200--215},
  organization = {Springer}
}

@article{chern2024can,
  title        = {{Can large language models be trusted for evaluation? scalable meta-evaluation of llms as evaluators via agent debate}},
  author       = {Chern, Steffi and Chern, Ethan and Neubig, Graham and others},
  year         = {2024},
  journal      = {arXiv preprint arXiv:2401.16788}
}

@article{garg2025rise,
  title        = {{The Rise of Small Language Models in Healthcare: A Comprehensive Survey}},
  author       = {Garg, Muskan and Raza, Shaina and Rayana, Shebuti and others},
  year         = {2025},
  journal      = {arXiv preprint arXiv:2504.17119}
}

@article{gilardi2023chatgpt,
  title        = {{ChatGPT outperforms crowd workers for text-annotation tasks}},
  author       = {Gilardi, Fabrizio and Alizadeh, Meysam and Kubli, Ma{\"e}l},
  year         = {2023},
  journal      = {Proceedings of the National Academy of Sciences},
  publisher    = {National Academy of Sciences},
  volume       = {120},
  number       = {30},
  pages        = {e2305016120}
}

@article{grattafiori2024llama,
  title        = {{The llama 3 herd of models}},
  author       = {Grattafiori, Aaron and Dubey, Abhimanyu and Jauhri, Abhinav and others},
  year         = {2024},
  journal      = {arXiv preprint arXiv:2407.21783}
}

@article{hassan2024automated,
  title        = {{Automated multi-label annotation for mental health illnesses using large language models}},
  author       = {Hassan, Abdelrahaman A and Hanafy, Radwa J and Fouda, Mohammed E},
  year         = {2024},
  journal      = {arXiv preprint arXiv:2412.03796}
}

@inproceedings{he2024annollm,
  title        = {{AnnoLLM: Making large language models to be better crowdsourced annotators}},
  author       = {He, Xingwei and Lin, Zhenghao and Gong, Yeyun and others},
  year         = {2024},
  booktitle    = {Proceedings of the 2024 Conference of the North American Chapter of the Association for Computational Linguistics: Human Language Technologies (Volume 6: Industry Track)},
}

@article{HERHAUSEN2025115491,
  title        = {{From words to insights: Text analysis in business research}},
  author       = {Dennis Herhausen and Stephan Ludwig and Ehsan Abedin and others},
  year         = {2025},
  journal      = {Journal of Business Research},
  volume       = {198},
  pages        = {115491},
  doi          = {https://doi.org/10.1016/j.jbusres.2025.115491},
  issn         = {0148-2963},
  url          = {https://www.sciencedirect.com/science/article/pii/S0148296325003145},
}

@inproceedings{holtzman2019curious,
    title={{The Curious Case of Neural Text Degeneration}},
    author={Ari Holtzman and Jan Buys and Li Du and Maxwell Forbes and Yejin Choi},
    booktitle={International Conference on Learning Representations},
    year={2020},
    url={https://openreview.net/forum?id=rygGQyrFvH}
}

@inproceedings{imdb,
  title        = {{Learning Word Vectors for Sentiment Analysis}},
  author       = {Maas, Andrew L.  and  Daly, Raymond E.  and  Pham, Peter T.  and others},
  year         = {2011},
  month        = jun,
  booktitle    = {Proceedings of the 49th Annual Meeting of the Association for Computational Linguistics: Human Language Technologies},
  publisher    = {Association for Computational Linguistics},
  address      = {Portland, Oregon, USA},
  pages        = {142--150},
  url          = {http://www.aclweb.org/anthology/P11-1015}
}

@inproceedings{khattab2024dspy,
  title        = {{DSPy: Compiling Declarative Language Model Calls into Self-Improving Pipelines}},
  author       = {Khattab, Omar and Singhvi, Arnav and Maheshwari, Paridhi and others},
  year         = {2024},
  journal      = {The Twelfth International Conference on Learning Representations}
}

@article{kojima2022large,
  title        = {{Large language models are zero-shot reasoners}},
  author       = {Kojima, Takeshi and Gu, Shixiang Shane and Reid, Machel and others},
  year         = {2022},
  journal      = {Advances in neural information processing systems},
  volume       = {35},
  pages        = {22199--22213}
}

@inproceedings{kwon2023efficient,
  title        = {{Efficient Memory Management for Large Language Model Serving with PagedAttention}},
  author       = {Woosuk Kwon and Zhuohan Li and Siyuan Zhuang and others},
  year         = {2023},
  booktitle    = {Proceedings of the ACM SIGOPS 29th Symposium on Operating Systems Principles}
}

@article{lee2025feedback,
  title        = {{Feedback descent: Open-ended text optimization via pairwise comparison}},
  author       = {Lee, Yoonho and Boen, Joseph and Finn, Chelsea},
  year         = {2025},
  journal      = {arXiv preprint arXiv:2511.07919}
}

@article{li2024cancerllm,
  title        = {{CancerLLM: A large language model in cancer domain}},
  author       = {Li, Mingchen and Huang, Jiatan and Yeung, Jeremy and others},
  year         = {2024},
  journal      = {arXiv preprint arXiv:2406.10459}
}

@inproceedings{lin2023toxicchat,
    title = "{T}oxic{C}hat: Unveiling Hidden Challenges of Toxicity Detection in Real-World User-{AI} Conversation",
    author = "Lin, Zi  and
      Wang, Zihan  and
      Tong, Yongqi  and
      Wang, Yangkun  and
      Guo, Yuxin  and
      Wang, Yujia  and
      Shang, Jingbo",
    editor = "Bouamor, Houda  and
      Pino, Juan  and
      Bali, Kalika",
    booktitle = "Findings of the Association for Computational Linguistics: EMNLP 2023",
    month = dec,
    year = "2023",
    address = "Singapore",
    publisher = "Association for Computational Linguistics",
    url = "https://aclanthology.org/2023.findings-emnlp.311/",
    doi = "10.18653/v1/2023.findings-emnlp.311",
    pages = "4694--4702",
}

@article{liu2025enhanced,
  title        = {{Enhanced large language models for effective screening of depression and anxiety}},
  author       = {Liu, June M and Gao, Mengxia and Sabour, Sahand and others},
  year         = {2025},
  journal      = {Communications Medicine},
  publisher    = {Nature Publishing Group UK London},
  volume       = {5},
  number       = {1},
  pages        = {457}
}

@article{lundberg2017unified,
  title        = {{A unified approach to interpreting model predictions}},
  author       = {Lundberg, Scott M and Lee, Su-In},
  year         = {2017},
  journal      = {Advances in neural information processing systems},
  volume       = {30}
}

@article{meyerson2025position,
  title        = {{Position: Scaling llm agents requires asymptotic analysis with llm primitives}},
  author       = {Meyerson, Elliot and Qiu, Xin},
  year         = {2025},
  journal      = {arXiv preprint arXiv:2502.04358}
}

@book{molnar2020interpretable,
  title        = {{Interpretable machine learning}},
  author       = {Molnar, Christoph},
  year         = {2020},
  publisher    = {Lulu. com}
}

@inproceedings{niculae-etal-2025-dr,
  title        = {{Dr. Copilot: A Multi-Agent Prompt Optimized Assistant for Improving Patient-Doctor Communication in {R}omanian}},
  author       = {Niculae, Andrei  and Cosma, Adrian  and Dumitrache, Cosmin  and Radoi, Emilian},
  year         = {2025},
  month        = nov,
  booktitle    = {Proceedings of the 2025 Conference on Empirical Methods in Natural Language Processing: Industry Track},
  publisher    = {Association for Computational Linguistics},
  address      = {Suzhou (China)},
  pages        = {1780--1792},
  doi          = {10.18653/v1/2025.emnlp-industry.125},
  isbn         = {979-8-89176-333-3},
  url          = {https://aclanthology.org/2025.emnlp-industry.125/},
}

@inproceedings{opsahl2024optimizing,
    title = "Optimizing Instructions and Demonstrations for Multi-Stage Language Model Programs",
    author = "Opsahl-Ong, Krista  and
      Ryan, Michael J  and
      Purtell, Josh  and
      Broman, David  and
      Potts, Christopher  and
      Zaharia, Matei  and
      Khattab, Omar",
    editor = "Al-Onaizan, Yaser  and
      Bansal, Mohit  and
      Chen, Yun-Nung",
    booktitle = "Proceedings of the 2024 Conference on Empirical Methods in Natural Language Processing",
    month = nov,
    year = "2024",
    address = "Miami, Florida, USA",
    publisher = "Association for Computational Linguistics",
    url = "https://aclanthology.org/2024.emnlp-main.525/",
    doi = "10.18653/v1/2024.emnlp-main.525",
    pages = "9340--9366",
}

@inproceedings{optuna,
  title        = {{Optuna: A Next-generation Hyperparameter Optimization Framework}},
  author       = {Akiba, Takuya and Sano, Shotaro and Yanase, Toshihiko and others},
  year         = {2019},
  booktitle    = {Proceedings of the 25th ACM SIGKDD International Conference on Knowledge Discovery \& Data Mining},
  location     = {Anchorage, AK, USA},
  publisher    = {Association for Computing Machinery},
  address      = {New York, NY, USA},
  series       = {KDD '19},
  pages        = {2623–2631},
  doi          = {10.1145/3292500.3330701},
  isbn         = {9781450362016},
  url          = {https://doi.org/10.1145/3292500.3330701},
  numpages     = {9}
}

@misc{outlines,
  title        = {{Efficient Guided Generation for Large Language Models}},
  author       = {Brandon T. Willard and Rémi Louf},
  year         = {2023},
  url          = {https://arxiv.org/abs/2307.09702},
  eprint       = {2307.09702},
  archiveprefix = {arXiv},
  primaryclass = {cs.CL}
}

@inproceedings{pryzant-etal-2023-automatic,
  title        = {{Automatic Prompt Optimization with ``Gradient Descent'' and Beam Search}},
  author       = {Pryzant, Reid  and Iter, Dan  and Li, Jerry  and Lee, Yin  and Zhu, Chenguang  and Zeng, Michael},
  year         = {2023},
  month        = dec,
  booktitle    = {Proceedings of the 2023 Conference on Empirical Methods in Natural Language Processing},
  publisher    = {Association for Computational Linguistics},
  address      = {Singapore},
  pages        = {7957--7968},
  doi          = {10.18653/v1/2023.emnlp-main.494},
  url          = {https://aclanthology.org/2023.emnlp-main.494/},
  editor       = {Bouamor, Houda  and Pino, Juan  and Bali, Kalika},
}

@article{reiss2023testing,
  title        = {{Testing the reliability of ChatGPT for text annotation and classification: A cautionary remark}},
  author       = {Reiss, Michael V},
  year         = {2023},
  journal      = {arXiv preprint arXiv:2304.11085}
}

@article{sinha2024pae,
  title        = {{Pae: Llm-based product attribute extraction for e-commerce fashion trends}},
  author       = {Sinha, Apurva and Gujral, Ekta},
  year         = {2024},
  journal      = {arXiv preprint arXiv:2405.17533}
}

@article{team2025gemma,
  title        = {{Gemma 3 technical report}},
  author       = {{Gemma Team} and Kamath, Aishwarya and Ferret, Johan and others},
  year         = {2025},
  journal      = {arXiv preprint arXiv:2503.19786}
}

@inproceedings{tigunova2019listening,
  title        = {{Listening between the lines: Learning personal attributes from conversations}},
  author       = {Tigunova, Anna and Yates, Andrew and Mirza, Paramita and others},
  year         = {2019},
  booktitle    = {The World Wide Web Conference},
  pages        = {1818--1828}
}

@misc{törnberg2023chatgpt4outperformsexpertscrowd,
  title        = {{ChatGPT-4 Outperforms Experts and Crowd Workers in Annotating Political Twitter Messages with Zero-Shot Learning}},
  author       = {Petter Törnberg},
  year         = {2023},
  url          = {https://arxiv.org/abs/2304.06588},
  eprint       = {2304.06588},
  archiveprefix = {arXiv},
  primaryclass = {cs.CL}
}

@article{wang2025slot,
  title        = {{SLOT: Structuring the Output of Large Language Models}},
  author       = {Wang, Darren Yow-Bang and Shen, Zhengyuan and Mishra, Soumya Smruti and others},
  year         = {2025},
  journal      = {arXiv preprint arXiv:2505.04016}
}

@article{xu2025metatextgrad,
  title        = {{metaTextGrad: Automatically optimizing language model optimizers}},
  author       = {Xu, Guowei and Yuksekgonul, Mert and Guestrin, Carlos and others},
  year         = {2025},
  journal      = {arXiv preprint arXiv:2505.18524}
}

@inproceedings{yang2023large,
  title        = {{Large language models as optimizers}},
  author       = {Yang, Chengrun and Wang, Xuezhi and Lu, Yifeng and others},
  year         = {2023},
  booktitle    = {The Twelfth International Conference on Learning Representations}
}

@article{yang2025qwen3,
  title        = {{Qwen3 technical report}},
  author       = {Yang, An and Li, Anfeng and Yang, Baosong and others},
  year         = {2025},
  journal      = {arXiv preprint arXiv:2505.09388}
}

@article{yuksekgonul2025optimizing,
  title        = {{Optimizing generative AI by backpropagating language model feedback}},
  author       = {Yuksekgonul, Mert and Bianchi, Federico and Boen, Joseph and others},
  year         = {2025},
  journal      = {Nature},
  volume       = {639},
  pages        = {609--616}
}

@misc{zhang2025dynamicadaptivefeaturegeneration,
  title        = {{Dynamic and Adaptive Feature Generation with LLM}},
  author       = {Xinhao Zhang and Jinghan Zhang and Banafsheh Rekabdar and others},
  year         = {2025},
  url          = {https://arxiv.org/abs/2406.03505},
  eprint       = {2406.03505},
  archiveprefix = {arXiv},
  primaryclass = {cs.LG}
}

@inproceedings{zhou-etal-2024-llm-feature,
  title        = {{An {LLM} Feature-based Framework for Dialogue Constructiveness Assessment}},
  author       = {Zhou, Lexin  and Farag, Youmna  and Vlachos, Andreas},
  year         = {2024},
  month        = nov,
  booktitle    = {Proceedings of the 2024 Conference on Empirical Methods in Natural Language Processing},
  publisher    = {Association for Computational Linguistics},
  address      = {Miami, Florida, USA},
  pages        = {5389--5409},
  doi          = {10.18653/v1/2024.emnlp-main.308},
  url          = {https://aclanthology.org/2024.emnlp-main.308/},
  editor       = {Al-Onaizan, Yaser  and Bansal, Mohit  and Chen, Yun-Nung}
}

@inproceedings{zhou2022large,
  title        = {{Large language models are human-level prompt engineers}},
  author       = {Zhou, Yongchao and Muresanu, Andrei Ioan and Han, Ziwen and others},
  year         = {2022},
  booktitle    = {The eleventh international conference on learning representations}
}

@article{zhou2025multi,
  title        = {{Multi-agent design: Optimizing agents with better prompts and topologies}},
  author       = {Zhou, Han and Wan, Xingchen and Sun, Ruoxi and others},
  year         = {2025},
  journal      = {arXiv preprint arXiv:2502.02533}
}

@article{zoller2021benchmark,
  title        = {{Benchmark and survey of automated machine learning frameworks}},
  author       = {Z{\"o}ller, Marc-Andr{\'e} and Huber, Marco F},
  year         = {2021},
  journal      = {Journal of artificial intelligence research},
  volume       = {70},
  pages        = {409--472}
}

@article{zytek2022need,
  title        = {{The need for interpretable features: Motivation and taxonomy}},
  author       = {Zytek, Alexandra and Arnaldo, Ignacio and Liu, Dongyu and others},
  year         = {2022},
  journal      = {ACM SIGKDD Explorations Newsletter},
  publisher    = {ACM New York, NY, USA},
  volume       = {24},
  number       = {1},
  pages        = {1--13}
}
